\documentclass[11pt,a4paper]{article}
\usepackage[hyperref]{naaclhlt2019}
\usepackage{times}
\usepackage{latexsym}
\usepackage{booktabs}
\usepackage{multirow}
\usepackage{pgfplots}
\usepackage{url}
\usepackage{tikz-dependency}
\usetikzlibrary{patterns,calc,shapes,backgrounds,shadows,positioning,fit,matrix,shapes.geometric}
\usetikzlibrary{automata, arrows}

\aclfinalcopy 
\def\aclpaperid{4055} 

\title{Practical Semantic Parsing for Spoken Language Understanding}

\author{Marco Damonte\thanks{\hspace{0.04cm} Work conducted while interning at Amazon Alexa AI.}\hspace{1cm}\\
  University of Edinburgh\\
  { \texttt{m.damonte@sms.ed.ac.uk}} \\\And
  Rahul Goel \quad Tagyoung Chung \\
  Amazon Alexa AI \\
  {\tt \{goerahul,tagyoung\}@amazon.com}}

\date{}

\begin{document}
\maketitle
\begin{abstract}
Executable semantic parsing is the task of converting natural language utterances into logical forms that can be directly used as queries to get a response. We build a transfer learning framework for executable semantic parsing. We show that the framework is effective for Question Answering (Q\&A) as well as for Spoken Language Understanding (SLU). We further investigate the case where a parser on a new domain can be learned by exploiting data on other domains, either via multi-task learning between the target domain and an auxiliary domain or via pre-training on the auxiliary domain and fine-tuning on the target domain. With either flavor of transfer learning, we are able to improve performance on most domains; we experiment with public data sets such as Overnight and NLmaps as well as with commercial SLU data. The experiments carried out on data sets that are different in nature show how executable semantic parsing can unify different areas of NLP such as Q\&A and SLU.
\end{abstract}


\section{Introduction}



Due to recent advances in speech recognition and language understanding, conversational interfaces such as Alexa, Cortana, and Siri are becoming more common. They currently have two large uses cases. First, a user can use them to complete a specific task, such as playing music. Second, a user can use them to ask questions where the questions are answered by querying knowledge graph or database back-end. Typically, under a common interface, there exist two disparate systems that can handle each use cases. The system underlying the first use case is known as a spoken language understanding (SLU) system. Typical commercial SLU systems rely on predicting a coarse user intent and then tagging each word in the utterance to the intent's slots. This architecture is popular due to its simplicity and robustness. On the other hand, Q\&A, which need systems to produce more complex structures such as trees and graphs, requires a more comprehensive understanding of human language.

One possible system that can handle such a task is an executable semantic parser~\cite{liang2013lambda,kate2005learning}. Given a user utterance, an executable semantic parser can generate tree or graph structures that represent logical forms that can be used to query a knowledge base or database. In this work, we propose executable semantic parsing as a common framework for both uses cases by framing SLU as executable semantic parsing that unifies the two use cases. For Q\&A, the input utterances are parsed into logical forms that represent the machine-readable representation of the question, while in SLU, they represent the machine-readable representation of the user intent and slots. One added advantage of using parsing for SLU is the ability to handle more complex linguistic phenomena such as coordinated intents that traditional SLU systems struggle to handle~\cite{agarwal2018parsing}. Our parsing model is an extension of the neural transition-based parser of \newcite{cheng2017learning}.


A major issue with semantic parsing is the availability of the annotated logical forms to train the parsers, which are expensive to obtain. A solution is to rely more on distant supervisions such as by using question--answer pairs~\cite{clarke2010driving,liang2013learning}. Alternatively, it is possible to exploit annotated logical forms from a different domain or related data set. In this paper, we focus on the scenario where data sets for several domains exist but only very little data for a new one is available and apply transfer learning techniques to it. A common way to implement transfer learning is by first pre-training the model on a domain on which a large data set is available and subsequently fine-tuning the model on the target domain \cite{thrun1996learning,zoph2016transfer}. We also consider a multi-task learning (MTL) approach. MTL refers to machine learning models that improve generalization by training on more than one task. MTL has been used for a number of NLP problems such as tagging \cite{collobert2008unified}, syntactic parsing \cite{luong2015multi}, machine translation \cite{dong2015multi,luong2015multi} and semantic parsing \cite{fan2017transfer}. See \newcite{caruana1997multitask} and \newcite{ruder2017overview} for an overview of MTL.


A good Q\&A data set for our domain adaptation scenario is the Overnight data set \cite{wang2015building}, which contains sentences annotated with Lambda Dependency-Based Compositional Semantics (Lambda DCS;~\citealt{liang2013lambda}) for eight different domains. However, it includes only a few hundred sentences for each domain, and its vocabularies are relatively small. We also experiment with a larger semantic parsing data set (NLmaps;~\citealt{lawrence2016nlmaps}). For SLU, we work with data from a commercial conversational assistant that has a much larger vocabulary size. One common issue in parsing is how to deal with rare or unknown words, which is usually addressed by either delexicalization or by implementing a copy mechanism \cite{gulcehre2016pointing}. We show clear differences in the outcome of these and other techniques when applied to data sets of varying sizes. 
Our contributions are as follows:
\begin{itemize}
\itemsep0em
\item We propose a common semantic parsing framework for Q\&A and SLU and demonstrate its broad applicability and effectiveness.
\item We report parsing baselines for Overnight for which exact match parsing scores have not been yet published.
\item We show that SLU greatly benefits from a copy mechanism, which is also beneficial for NLmaps but not Overnight. 
\item We investigate the use of transfer learning and show that it can facilitate parsing on low-resource domains.
\end{itemize}




\section{Transition-based Parser}
Transition-based parsers are widely used for dependency parsing \cite{nivre2008algorithms,dyer2015transition} and they have been also applied to semantic parsing tasks \cite{wang2015transition, cheng2017learning}. 

In syntactic parsing, a transition system is usually defined as a quadruple: $T = \{S, A, I, E\}$, where $S$ is a set of states, $A$ is a set of actions, $I$ is the initial state, and $E$ is a set of end states. A state is composed of a buffer, a stack, and a set of arcs: $S = (\beta, \sigma, A)$. In the initial state, the buffer contains all the words in the input sentence while the stack and the set of subtrees are empty: $S_0 = (w_0|\dots|w_N, \emptyset, \emptyset)$. Terminal states have empty stack and buffer: $S_T = (\emptyset, \emptyset, A)$. 

During parsing, the stack stores words that have been removed from the buffer but have not been fully processed yet. Actions can be performed to advance the transition system's state: they can either consume words in the buffer and move them to the stack (\textsc{SHIFT}) or combine words in the stack to create new arcs (\textsc{LEFT-ARC} and \textsc{RIGHT-ARC}, depending on the direction of the arc)\footnote{There are multiple different transition systems. The example we describe here is that of {\it arc-standard} system \cite{nivre2004incrementality} for projective dependency parsing.}. Words in the buffer are processed left-to-right until an end state is reached, at which point the set of arcs will contain the full output tree.

The parser needs to be able to predict the next action based on its current state. Traditionally, supervised techniques are used to learn such classifiers, using a parallel corpus of sentences and their output trees. Trees can be converted to states and actions using an oracle system. For a detailed explanation of transition-based parsing, see \newcite{nivre2003efficient} and \newcite{nivre2008algorithms}.

\subsection{Neural Transition-based Parser with Stack-LSTMs}
\label{sec:baseline}

In this paper, we consider the neural executable semantic parser of \newcite{cheng2017learning}, which follows the transition-based parsing paradigm. Its transition system differs from traditional systems as the words are not consumed from the buffer because
in executable semantic parsing, there are no strict alignments between words in the input and nodes in the tree. The neural architecture encodes the buffer using a Bi-LSTM \cite{graves2012supervised} and the stack as a Stack-LSTM \cite{dyer2015transition}, a recurrent network that allows for push and pop operations.
Additionally, the previous actions are also represented with an LSTM. The output of these networks is fed into feed-forward layers and softmax layers are used to predict the next action given the current state. The possible actions are \textsc{REDUCE}, which pops an item from the stack, \textsc{TER}, which creates a terminal node (i.e., a leaf in the tree), and \textsc{NT}, which creates a non-terminal node. When the next action is either TER or NT, additional softmax layers predict the output token to be generated. Since the buffer does not change while parsing, an attention mechanism is used to focus on specific words given the current state of the parser.

We extend the model of \newcite{cheng2017learning} by adding character-level embeddings and a copy mechanism. When using only word embeddings, out-of-vocabulary words are usually mapped to one embedding vector and do not exploit morphological features. 
Our model encodes words by feeding each character embedding onto an LSTM and concatenate its output to the word embedding:
\begin{equation}
x = \{e_w; h_c^M\},
\end{equation}
where $e_w$ is the word embedding of the input word $w$ and $h_c^M$ is the last hidden state of the character-level LSTM over the characters of the input word $w = c_0, \dots, c_M$.

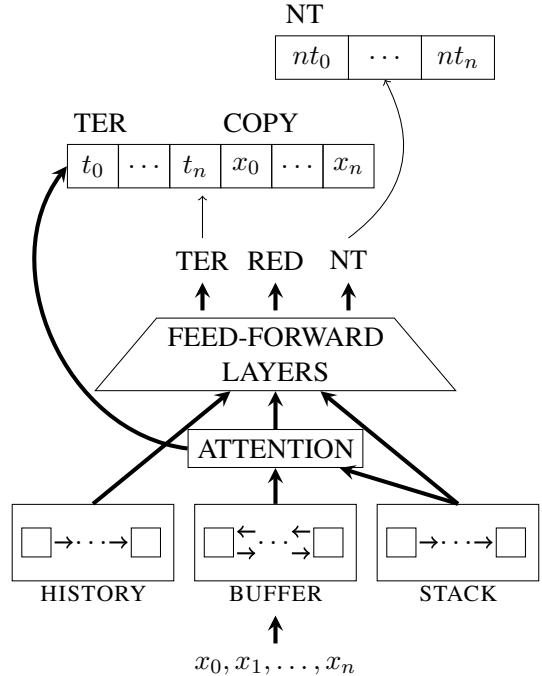
\begin{figure}[ht!]
 \centering
 \scalebox{0.96}{
  \begin{tikzpicture}
  \draw (2.5,-0.7) node(sent) [rectangle, minimum width=2.8cm, minimum height=0.6cm, rounded corners=0.1cm] {$x_0, x_1, \dots, x_n$};
  
  \draw (0, 0.3) node(hist_lab) [rectangle, minimum width=2.8cm, minimum height=0.6cm, rounded corners=0.1cm] {\small{\textsc{HISTORY}}};  
  \draw (0,1) node(hist) [draw, rectangle, minimum width=1cm, minimum height=0.2cm] {
      \begin{tikzpicture}  
          \draw (0,0) node(x1) [draw, rectangle, minimum width=0.4cm, minimum height=0.4cm, fill=white] {};
          \draw (0.75,0) node(x3) [minimum width=0.4cm, minimum height=0.8cm] {$\dots$};
          \draw (1.5,0) node(x4) [draw, rectangle, minimum width=0.4cm, minimum height=0.4cm, fill=white] {};
          \draw [->, thick] (0.25,0) -- (0.5,0) node {};
          \draw [->, thick] (1,0) -- (1.25,0) node {};
        \end{tikzpicture}    
  };
  
  \draw (2.5,0.3) node(buf_lab) [rectangle, minimum width=2.8cm, minimum height=0.6cm, rounded corners=0.1cm] {\small{\textsc{BUFFER}}};    
  \draw [->,>=stealth, ultra thick] (sent.north) -- (buf_lab.south) node {};
  \draw (2.5,1) node(buf) [draw, rectangle, minimum width=1cm, minimum height=0.2cm] {
      \begin{tikzpicture}  
          \draw (0,0) node(x1) [draw, rectangle, minimum width=0.4cm, minimum height=0.4cm, fill=white] {};
          \draw (0.75,0) node(x3) [minimum width=0.4cm, minimum height=0.8cm] {$\dots$};
          \draw (1.5,0) node(x4) [draw, rectangle, minimum width=0.4cm, minimum height=0.4cm, fill=white] {};
          \draw [->, thick] (0.25,-0.15) -- (0.5,-0.15) node {};
          \draw [->, thick] (1,-0.15) -- (1.25,-0.15) node {};
          \draw [<-, thick] (0.25,0.15) -- (0.5,0.15) node {};
          \draw [<-, thick] (1,0.15) -- (1.25,0.15) node {};          
        \end{tikzpicture}    
  };
  
  \draw (5,0.3) node(stack_lab) [rectangle, minimum width=2.8cm, minimum height=0.6cm, rounded corners=0.1cm] {\small{\textsc{STACK}}};      
  \draw (5,1) node(stack) [draw, rectangle, minimum width=1cm, minimum height=0.2cm] {
      \begin{tikzpicture}  
          \draw (0,0) node(x1) [draw, rectangle, minimum width=0.4cm, minimum height=0.4cm, fill=white] {};
          \draw (0.75,0) node(x3) [minimum width=0.4cm, minimum height=0.8cm] {$\dots$};
          \draw (1.5,0) node(x4) [draw, rectangle, minimum width=0.4cm, minimum height=0.4cm, fill=white] {};
          \draw [->, thick] (0.25,0) -- (0.5,0) node {};
          \draw [->, thick] (1,0) -- (1.25,0) node {};
        \end{tikzpicture}    
  };  
   
  \draw (2.5,2.3) node(attention) [draw, rectangle, minimum width=1cm, minimum height=0.2cm]{ATTENTION};
  
  \draw (2.5,3.6) node(feed) [draw, trapezium, trapezium angle=50, text width=3cm, minimum width=3cm, minimum height=0.6cm, align=center]{FEED-FORWARD LAYERS};
  
  \draw (1.5,4.9) node(s1) [rectangle, minimum width=1cm, minimum height=0.6cm]{TER};
  \draw (2.5,4.9) node(s2) [rectangle, minimum width=1cm, minimum height=0.6cm]{RED};
  \draw (3.5,4.9) node(s3) [rectangle, minimum width=1cm, minimum height=0.6cm]{NT};
  \draw (1.5,3.9) node(s11) [rectangle, minimum width=1cm, minimum height=0.6cm]{};
  \draw (2.5,3.9) node(s21) [rectangle, minimum width=1cm, minimum height=0.6cm]{};
  \draw (3.5,3.9) node(s31) [rectangle, minimum width=1cm, minimum height=0.6cm]{};  
  \draw (1.5,6.1) node(s12) [rectangle, minimum width=1cm, minimum height=0.6cm]{};
  
  \draw (0,6.2) node(ter) [draw, rectangle, minimum width=0.7cm, minimum height=0.6cm]{$t_0$};
  \draw (0.7,6.2) node(feed0) [draw, rectangle, minimum width=0.7cm, minimum height=0.6cm]{$\dots$};
  \draw (1.4,6.2) node(feed0) [draw, rectangle, minimum width=0.7cm, minimum height=0.6cm]{$t_n$};
  \draw (2.1,6.2) node(feed0) [draw, rectangle, minimum width=0.7cm, minimum height=0.6cm]{$x_0$};
  \draw (2.8,6.2) node(feed0) [draw, rectangle, minimum width=0.7cm, minimum height=0.6cm]{$\dots$};
  \draw (3.5,6.2) node(feed0) [draw, rectangle, minimum width=0.7cm, minimum height=0.6cm]{$x_n$};
  \draw (0.1,6.8) node(ter_lab) [rectangle, minimum width=1cm, minimum height=0.2cm]{TER};
  \draw (2.3,6.8) node(copy_lab) [rectangle, minimum width=1cm, minimum height=0.2cm]{COPY};
  
  \draw (3,7.7) node(nt0) [draw, rectangle, minimum width=1cm, minimum height=0.6cm]{$nt_0$};
  \draw (4,7.7) node(nt1) [draw, rectangle, minimum width=1cm, minimum height=0.6cm]{$\dots$};
  \draw (5,7.7) node(feed0) [draw, rectangle, minimum width=1cm, minimum height=0.6cm]{$nt_n$};  
  \draw (2.9,8.3) node(nt_lab) [rectangle, minimum width=1cm, minimum height=0.2cm]{NT};

  \draw [->,>=stealth, ultra thick] (s11.north) -- (s1.south); 
  \draw [->,>=stealth, ultra thick] (s21.north) -- (s2.south); 
  \draw [->,>=stealth, ultra thick] (s31.north) -- (s3.south); 

  \draw [->,>=stealth, ultra thick] (buf.north) -- (attention.south); 
  \draw [->,>=stealth, ultra thick] (stack.north) -- (attention); 
  \draw [->,>=stealth, ultra thick] (stack.north) -- (feed); 
  \draw [->,>=stealth, ultra thick] (hist.north) -- (feed);
  \draw [->,>=stealth, ultra thick] (attention.north) -- (feed);
  
  \draw [bend left=60,distance=1.7cm,->,>=stealth, ultra thick] (attention.west) to node [left] {} (ter.west);
  \draw [->] (s1.north) -- (s12.south);
  
  \draw [bend right=40,distance=1cm,->] (s3.north) to node [left] {} (nt1.south);  
  \end{tikzpicture}
  }
\caption{{\small The full neural transition-based parsing model. Representations of stack, buffer, and previous actions are used to predict the next action. When the TER or NT actions are chosen, further layers are used to predict (or copy) the token.}}
\label{fig:model}
\end{figure}

Rare words are usually handled by either delexicalizing
the output or by using a copy mechanism. Delexicalization involves substituting named entities with a specific token in an effort to reduce the number of rare and unknown words. Copy relies on the fact that when rare or unknown words must be generated, they usually appear in the same form in the input sentence and they can be therefore copied from the input itself. Our copy implementation follows the strategy of \citet{fan2017transfer}, where the output of the generation layer is concatenated to the scores of an attention mechanism~\cite{bahdanau2015neural}, which express the relevance of each input word with respect to the current state of the parser. In the experiments that follow, we compare delexicalization with copy mechanism on different setups. A depiction of the full model is shown in Figure~\ref{fig:model}.

\section{Transfer learning}
We consider the scenario where large training corpora are available for some domains and we want to bootstrap a parser for a new domain where little training data is available. We investigate the use of two transfer learning approaches: pre-training and multi-task learning. 



For MTL, the different tasks share most of the architecture and only the output layers, which are responsible for predicting the output tokens, are separate for each task. When multi-tasking across domains of the same data set, we expect that most layers of the neural parser, such as the ones responsible for learning the word embeddings and the stack and buffer representation, will learn similar features and can, therefore, be shared. We implement two different MTL setups: a) when separate heads are used for both the \textsc{TER} classifier and the \textsc{NT} classifier, which is expected to be effective when transferring across tasks that do not share output vocabulary; and b) when a separate head is used only for the \textsc{TER} classifier, more appropriate when the non-terminals space is mostly shared. 

\section{Data}
\label{sec:data}

In order to investigate the flexibility of the executable semantic parsing framework, we evaluate models on Q\&A data sets as well as on commercial SLU data sets. For Q\&A, we consider Overnight \cite{wang2015building} and NLmaps \cite{lawrence2016nlmaps}.

\paragraph{Overnight}
It contains sentences annotated with Lambda DCS \cite{liang2013lambda}. The sentences are divided into eight domains: \emph{calendar}, \emph{blocks}, \emph{housing}, \emph{restaurants}, \emph{publications}, \emph{recipes}, \emph{socialnetwork}, and \emph{basketball}. As shown in Table~\ref{tab:data}, the number of sentences and the terminal vocabularies are small, which makes the learning more challenging, preventing us from using data-hungry approaches such as sequence-to-sequence models. The current state-of-the-art results, to the best of our knowledge, are reported by \newcite{su2017crossdomain}. Previous work on this data set use denotation accuracy as a metric. In this paper, we use logical form exact match accuracy across all data sets.

\paragraph{NLmaps}
It contains more than two thousand questions about geographical facts, retrieved from OpenStreetMap \cite{haklay2008openstreetmap}. Unfortunately, this data set is not divided into sub-domains. While NLmaps has comparable sizes with some of the Overnight domains, its vocabularies are much larger: containing 160 terminals, 24 non-terminals and 280 word types (Table~\ref{tab:data}). The current state-of-the-art results on this data set are reported by \newcite{duong2017multilingual}.

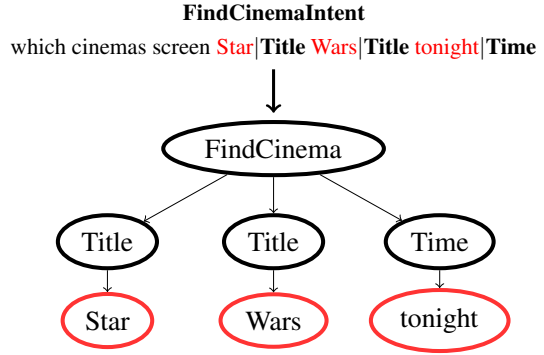
\begin{figure}[t]
 \centering
 \scalebox{0.95}{
  \begin{tikzpicture}
    \draw (0,6.8) node(r) [ellipse,draw,ultra thick] {FindCinema};
    \draw (2.3,5.5) node(a3) [ellipse,draw,ultra thick] {Time};
    \draw (0,5.5) node(a2) [ellipse,draw,ultra thick] {Title};
    \draw (-2.3,5.5) node(a1) [ellipse,draw,ultra thick] {Title};    
    \draw (2.3,4.4) node(sh) [ellipse,draw=red!80,ultra thick] {tonight};
    \draw (0,4.4) node(se) [ellipse,draw=red!80,ultra thick] {Wars};
    \draw (-2.3,4.4) node(la) [ellipse,draw=red!80,ultra thick] {Star};
    \draw (0,8.7) node(z1) [] {\small{\textbf{FindCinemaIntent}}};
    \draw (0,8.2) node(z2) [] {\small{which cinemas screen {\color{red}Star}$\vert$\textbf{Title} {\color{red}Wars}$\vert$\textbf{Title} {\color{red}tonight}$\vert$\textbf{Time}}};
  
       \draw [->,very thick] (z2) -- node[]{} (0,7.3);
     \draw [->] (r) -- node[]{} (a1);
     \draw [->] (r) -- node[]{} (a2);
     \draw [->] (r) -- node[]{} (a3);
     \draw [->] (a1) -- node[]{} (la);
     \draw [->] (a2) -- node[]{} (se);
     \draw [->] (a3) -- node[]{} (sh);     
     
  \end{tikzpicture}}
 \caption{{\small Conversion from intent/slot tags to tree for the sentence \emph{Which cinemas screen Star Wars tonight?}}}
 \label{fig:conversion}
\end{figure}

\paragraph{SLU}
We select five domains from our SLU data set: \emph{search}, \emph{recipes}, \emph{cinema}, \emph{bookings}, and \emph{closet}. In order to investigate the use case of a new low-resource domain exploiting a higher-resource domain, we selected a mix of high-resource and low-resource domains. Details are shown in Table~\ref{tab:data}. We extracted shallow trees from data originally collected for intent/slot tagging: intents become the root of the tree, slot types are attached to the roots as their children and slot values are in turn attached to their slot types as their children. An example is shown in Figure~\ref{fig:conversion}. A similar approach to transform intent/slot data into tree structures has been recently employed by \newcite{gupta}.

\begin{table}
\centering
\scalebox{0.95}{
\begin{tabular}{lrrrr}
\toprule
\textbf{DOMAIN} & \textbf{\#} & \textbf{TER} & \textbf{NT} & \textbf{Words}\\
\midrule
\multicolumn{5}{c}{Q\&A}\\
calendar & 535 & 31 & 13 & 114 \\
blocks & 1276 & 30 & 13 & 99 \\
housing & 601 & 34 & 13 & 109 \\
restaurants & 1060 & 40 & 13 & 144 \\
publications & 512 & 24 & 12 & 80 \\
recipes & 691 & 30 & 13 & 121 \\
social & 2828 & 56 & 16 & 225 \\
basketball & 1248 & 40 & 15 & 148 \\
\midrule
NLmaps & 1200 & 160 & 24 & 280 \\
\midrule
\multicolumn{5}{c}{SLU}\\
search & 23706 & 1621 & 51 & 1780 \\
recipes & 18721 & 530 & 40 & 643 \\
cinema & 13180 & 806 & 36 & 923 \\
bookings & 1280 & 10 & 19 & 42 \\
closet & 943 & 63 & 13 & 107 \\
\midrule
\bottomrule
\end{tabular}}
\caption{{\small Details of training data. \# is the number of sentences, \emph{TER} is the terminal vocabulary size, \emph{NT} is the non-terminal vocabulary size and \emph{Words} is the input vocabulary size.}}
\label{tab:data}
\end{table}

\section{Experiments}

We first run experiments on single-task semantic parsing to observe the differences among the three different data sources discussed in Section~\ref{sec:data}. Specifically, we explore the impact of an attention mechanism on the performance as well as the comparison between delexicalization and a copy mechanism for dealing with data sparsity. The metric used to evaluate parsers is the exact match accuracy, defined as the ratio of sentences correctly parsed.

\subsection{Attention}
\label{sec:attention}

Because the buffer is not consumed as in traditional transition-based parsers, \newcite{cheng2017learning} use an additive attention mechanism \cite{bahdanau2015neural} to focus on the more relevant words in the buffer for the current state of the stack.



In order to find the impact of attention on the different data sets, we run ablation experiments, as shown in Table~\ref{tab:comp} (left side). We found that attention between stack and buffer is not always beneficial: it appears to be helpful for larger data sets while harmful for smaller data sets. Attention is, however, useful for NLmaps, regardless of the data size. Even though NLmaps data is similarly sized to some of the Overnight domains, its terminal space is considerably larger, perhaps making attention more important even with a smaller data set. On the other hand, the high-resource SLU's \emph{cinema} domain is not able to benefit from the attention mechanism. We note that the performance of this model on NLmaps falls behind the state of the art \citep{duong2017multilingual}. The hyper-parameters of our model were however not tuned on this data set.

\begin{table}
\centering
\scalebox{0.95}{
\begin{tabular}{lcc|cc}
\toprule
 \textbf{DOMAIN} & \textbf{BL} & \textbf{$-$Att} & \textbf{$+$Delex} & \textbf{$+$Copy}\\
\midrule
 calendar & 38.1 & \textbf{43.5} & 4.20 & 32.1\\ 
 blocks & 22.6 & \textbf{25.1} & 24.3 & 22.8 \\
 housing & 19.0 & \textbf{29.6} & 6.90 & 21.2\\
 restaurants & 32.2 & \textbf{37.3} & 21.7 & 33.7\\
 publications & 27.3 & \textbf{32.9} & 11.8 & 26.1\\
 recipes & 47.7 & \textbf{58.3} & 24.1 & 48.1\\
 social & 44.9 & \textbf{51.2} & 47.7 & 50.9\\
 basketball & 65.2 & \textbf{69.6} & 38.6 & 66.5\\
\midrule
NLmaps  & 44.9 & 43.5 & 46.4 & \textbf{60.7}\\
\midrule
 search & 35.6 & 34.9 & 29.2 & \textbf{52.7} \\
 recipes & 40.9 & 37.9 & 37.7 & \textbf{47.6}\\
 cinema & 31.5 & 35.5 & 35.7 & \textbf{56.9}\\
 bookings & 72.3 & 77.7& 72.3 & \textbf{77.7}\\
 closet & 17.6 & 35.9& 29.2 & \textbf{44.1}\\
\midrule
\bottomrule
\end{tabular}}
\caption{{\small Left side: Ablation experiments on attention mechanism. Right side: Comparison between delexicalization and copy mechanism. \emph{BL} is the model of Section~\ref{sec:baseline}, \emph{$-$Att} refers to the same model without attention, \emph{$+$Delex} is the system with delexicalization and in \emph{$+$Copy} we use a copy mechanism instead. The scores indicate the percentage of correct parses.}}
\label{tab:comp}
\end{table}

\subsection{Handling Sparsity}

A popular way to deal with the data sparsity problem is to delexicalize the data, that is replacing rare and unknown words with coarse categories. In our experiment, we use a named entity recognition system\footnote{https://spacy.io} to replace names with their named entity types. Alternatively, it is possible to use a copy mechanism to enable the decoder to copy rare words from the input rather than generating them from its limited vocabulary.

We compare the two solutions across all data sets on the right side of Table~\ref{tab:comp}. Regardless of the data set, the copy mechanism generally outperforms delexicalization. We also note that delexicalization has unexpected catastrophic effects on exact match accuracy for \emph{calendar} and \emph{housing}. For Overnight, however, the system with copy mechanism is outperformed by the system without attention. This is unsurprising as the copy mechanism is based on attention, which is not effective on Overnight (Section~\ref{sec:attention}). The inefficacy of copy mechanisms on the Overnight data set was also discussed in \newcite{jia2016data}, where answer accuracy, rather than parsing accuracy, was used as a metric. As such, the results are not directly comparable.  

For NLmaps and all SLU domains, using a copy mechanism results in an average accuracy improvement of 16\% over the baseline. It is worth noting that the copy mechanism is unsurprisingly effective for SLU data due to the nature of the data set: the SLU trees were obtained from data collected for slot tagging, and as such, each leaf in the tree has to be copied from the input sentence.


Even though Overnight often yields different conclusions, most likely due to its small vocabulary size, the similar behaviors observed for NLmaps and SLU is reassuring, confirming that it is possible to unify Q\&A and SLU under the same umbrella framework of executable semantic parsing. 

In order to compare the NLmaps results with \newcite{lawrence2016nlmaps}, we also compute F1 scores for the data set. Our baseline outperforms previous results, achieving a score of 0.846. Our best F1 results are also obtained when adding the copy mechanism, achieving a score of 0.874\@.


\subsection{Transfer Learning}
The first set of experiments involve transfer learning across Overnight domains. For this data set, the non-terminal vocabulary is mostly shared across domains. As such, we use the architecture where only the TER output classifier is not shared. 
Selecting the best auxiliary domain by maximizing the overlap with the main domain was not successful, and we instead performed an exhaustive search over the domain pairs on the development set. In the interest of space, for each main domain, we report results for the best auxiliary domain (Table~\ref{tab:mtl_overnight}). 
 We note that MTL and pre-training provide similar results and provide an average improvement of 4\%\@. As expected, we observe more substantial improvements for smaller domains. 

\begin{table}
\centering
\scalebox{0.95}{
\begin{tabular}{llcc}
\toprule
\textbf{DOMAIN} & \textbf{BL} & \textbf{MTL} & \textbf{PRETR.}\\
\midrule
calendar & 43.5 & \textbf{48.8} & 48.2\\ 
blocks & 25.1 & 24.1 & \textbf{25.1}\\ 
housing & 29.6 & 38.1 & \textbf{38.1}\\ 
restaurants & 37.3 & \textbf{39.2} & 36.7\\ 
publications & 32.9 & 37.3 & \textbf{40.4}\\ 
recipes & 58.3 & \textbf{63.4} & 63.0\\ 
social & 51.2 & 52.4 & \textbf{54.5}\\ 
basketball & 69.6 & 69.1 & \textbf{71.1}\\  
\bottomrule
\end{tabular}}
\caption{{\small Transfer learning results for the Overnight domains. \emph{BL $-$ Att} is the model without transfer learning. \emph{PRETR.} stands for pre-training. Again, we report exact match accuracy.}}
\label{tab:mtl_overnight}
\end{table}

We performed the same set of experiments on the SLU domains, as shown in Table~\ref{tab:mtl_slu}. In this case, the non-terminal vocabulary can vary significantly across domains. We therefore choose to use the MTL architecture where both TER and NT output classifiers are not shared. Also for SLU, there is no clear winner between pre-training and MTL. Nevertheless, they always outperform the baseline, demonstrating the importance of transfer learning, especially for smaller domains.

\begin{table}
\centering
\scalebox{0.95}{
\begin{tabular}{llcc}
\toprule
\textbf{DOMAIN} & \textbf{BL $+$ Copy} & \textbf{MTL} & \textbf{PRETR.}\\
\midrule
search & 52.7 & 52.3 & \textbf{53.1} \\ 
cinema & 56.9 & \textbf{57.7} & 56.4 \\ 
bookings & 77.7 & \textbf{81.2} & 78.0 \\ 
closet & 44.1 & \textbf{52.5} & 50.8 \\ 
\bottomrule
\end{tabular}}
\caption{{\small Transfer learning results for SLU domains. \emph{BL $+$ Copy} is the model without transfer learning. \emph{PRETR.} stands for pre-training. Again, the numbers are exact match accuracy.}}
\label{tab:mtl_slu}
\end{table}\textbf{}

While the focus of this transfer learning framework is in exploiting high-resource domains annotated in the same way as a new low-resource domain, we also report a preliminary experiment on transfer learning across tasks. We selected the \emph{recipes} domain, which exists in both Overnight and SLU. While the SLU data set is significantly different from Overnight, deriving from a corpus annotated with intent/slot labels, as discussed in Section~\ref{sec:data}, we found promising results using pre-training, increasing the accuracy from 58.3 to 61.1\@. A full investigation of transfer learning across domains belonging to heterogeneous data sets is left for future work.

The experiments on transfer learning demonstrate how parsing accuracy on low-resource domains can be improved by exploiting other domains or data sets. Except for the Overnight's \emph{blocks} domain, which is one of the largest in Overnight, all domains in Overnight and SLU were shown to provide better results when either MTL or pre-training was used, with the most significant improvements observed for low-resource domains.

\section{Related work}

A large collection of logical forms of different nature exist in the semantic parsing literature: semantic role schemes \cite{palmer2005proposition,meyers2004annotating,baker1998berkeley}, syntax/semantics interfaces \cite{steedman1996surface}, executable logical forms \cite{liang2013lambda,kate2005learning}, and general purpose meaning representations \cite{Banarescu13abstractmeaning,abend2013universal}. We adopt executable logical forms in this paper. The Overnight data set
uses Lambda DCS,
the NLmaps data set extracts meaning representations from OpenStreetMap, and the SLU data set contains logical forms reminiscent of Lambda DCS that can be used to perform actions and query databases. State-of-the-art results are reported in \newcite{su2017crossdomain} for Overnight and \newcite{duong2017multilingual} for NLmaps.\footnote{The results on Overnight are not computed on the logical form they produce but on the answer they obtain using the logical form as a query. As such, their results are not directly comparable to ours.}

Our semantic parsing model is an extension of the executable semantic parser of \newcite{cheng2017learning}, which is inspired by Recurrent Neural Network Grammars \cite{dyer2016recurrent}. We extend the model with ideas inspired by \citet{gulcehre2016pointing} and \citet{luong2016achieving}.

We build our multi-task learning architecture upon the rich literature on the topic. MTL was first introduce in \newcite{caruana1997multitask}. It has been since used for a number of NLP problems such as tagging \cite{collobert2008unified}, syntactic parsing \cite{luong2015multi}, and machine translation \cite{dong2015multi,luong2015multi}. The closest to our work is \newcite{fan2017transfer}, where MTL architectures are built on top of an attentive sequence-to-sequence model \cite{bahdanau2015neural}.
We instead focus on transfer learning across domains of the same data sets and employ a different architecture which promises to be less data-hungry than sequence-to-sequence models.

Typical SLU systems
rely on domain-specific semantic parsers that identify intents and
slots in a sentence. Traditionally, these tasks were performed by
linear machine learning models \cite{sha2003shallow} but more recently jointly-trained DNN
models are used \cite{mesnil2015using,hakkani2016multi} with differing contexts~\cite{gupta2018efficient,naik2018modal}. More recently there has been work on extending the traditional intent/slot framework using targeted parsing to handle more complex linguistic phenomenon like coordination \cite{gupta2018semantic,agarwal2018parsing}. 


\section{Conclusions}

We framed SLU as an executable semantic parsing task, which addresses a limitation of current commercial SLU systems. By applying our framework to different data sets, we demonstrate that the framework is effective for Q\&A as well as for SLU. We explored a typical scenario where it is necessary to learn a semantic parser for a new domain with little data, but other high-resource domains are available. We show the effectiveness of our system and both pre-training and MTL on different domains and data sets. Preliminary experiment results on transfer learning across domains belonging to heterogeneous data sets suggest future work in this area.

\section*{Acknowledgments}
The authors would like to thank the three anonymous reviewers for their comments and the Amazon Alexa AI team members for their feedback.

\bibliography{reference}

\begin{thebibliography}{43}
\expandafter\ifx\csname natexlab\endcsname\relax\def\natexlab#1{#1}\fi

\bibitem[{Abend and Rappoport(2013)}]{abend2013universal}
Omri Abend and Ari Rappoport. 2013.
\newblock Universal conceptual cognitive annotation (ucca).
\newblock In \emph{Proceedings of ACL}.

\bibitem[{Agarwal et~al.(2018)Agarwal, Goel, Chung, Sethi, Mandal, and
  Matsoukas}]{agarwal2018parsing}
Sanchit Agarwal, Rahul Goel, Tagyoung Chung, Abhishek Sethi, Arindam Mandal,
  and Spyros Matsoukas. 2018.
\newblock Parsing coordination for spoken language understanding.
\newblock \emph{arXiv preprint arXiv:1810.11497}.

\bibitem[{Bahdanau et~al.(2015)Bahdanau, Cho, and Bengio}]{bahdanau2015neural}
Dzmitry Bahdanau, Kyunghyun Cho, and Yoshua Bengio. 2015.
\newblock Neural machine translation by jointly learning to align and
  translate.

\bibitem[{Baker et~al.(1998)Baker, Fillmore, and Lowe}]{baker1998berkeley}
Collin~F Baker, Charles~J Fillmore, and John~B Lowe. 1998.
\newblock The berkeley framenet project.
\newblock In \emph{Proceedings of COLING}.

\bibitem[{Banarescu et~al.(2013)Banarescu, Bonial, Cai, Georgescu, Griffitt,
  Hermjakob, Knight, Koehn, Palmer, and Schneider}]{Banarescu13abstractmeaning}
Laura Banarescu, Claire Bonial, Shu Cai, Madalina Georgescu, Kira Griffitt, Ulf
  Hermjakob, Kevin Knight, Philipp Koehn, Martha Palmer, and Nathan Schneider.
  2013.
\newblock Abstract meaning representation for sembanking.
\newblock \emph{Linguistic Annotation Workshop}.

\bibitem[{Caruana(1997)}]{caruana1997multitask}
Rich Caruana. 1997.
\newblock Multitask learning.
\newblock \emph{Machine learning}, 28(1):41--75.

\bibitem[{Cheng et~al.(2017)Cheng, Reddy, Saraswat, and
  Lapata}]{cheng2017learning}
Jianpeng Cheng, Siva Reddy, Vijay Saraswat, and Mirella Lapata. 2017.
\newblock Learning structured natural language representations for semantic
  parsing.
\newblock In \emph{Proceedings of the 55th Annual Meeting of the Association
  for Computational Linguistics (Volume 1: Long Papers)}, volume~1, pages
  44--55.

\bibitem[{Clarke et~al.(2010)Clarke, Goldwasser, Chang, and
  Roth}]{clarke2010driving}
James Clarke, Dan Goldwasser, Ming-Wei Chang, and Dan Roth. 2010.
\newblock Driving semantic parsing from the world's response.
\newblock In \emph{Proceedings of CoNLL}. Association for Computational
  Linguistics.

\bibitem[{Collobert and Weston(2008)}]{collobert2008unified}
Ronan Collobert and Jason Weston. 2008.
\newblock A unified architecture for natural language processing: Deep neural
  networks with multitask learning.
\newblock In \emph{Proceedings of ICML}.

\bibitem[{Dong et~al.(2015)Dong, Wu, He, Yu, and Wang}]{dong2015multi}
Daxiang Dong, Hua Wu, Wei He, Dianhai Yu, and Haifeng Wang. 2015.
\newblock Multi-task learning for multiple language translation.
\newblock In \emph{Proceedings of ACL}.

\bibitem[{Duong et~al.(2017)Duong, Afshar, Estival, Pink, Cohen, and
  Johnson}]{duong2017multilingual}
Long Duong, Hadi Afshar, Dominique Estival, Glen Pink, Philip Cohen, and Mark
  Johnson. 2017.
\newblock Multilingual semantic parsing and code-switching.
\newblock In \emph{Proceedings of CoNLL 2017}.

\bibitem[{Dyer et~al.(2015)Dyer, Ballesteros, Ling, Matthews, and
  Smith}]{dyer2015transition}
Chris Dyer, Miguel Ballesteros, Wang Ling, Austin Matthews, and Noah~A Smith.
  2015.
\newblock Transition-based dependency parsing with stack long short-term
  memory.
\newblock In \emph{Proceedings of ACL}.

\bibitem[{Dyer et~al.(2016)Dyer, Kuncoro, Ballesteros, and
  Smith}]{dyer2016recurrent}
Chris Dyer, Adhiguna Kuncoro, Miguel Ballesteros, and Noah~A Smith. 2016.
\newblock Recurrent neural network grammars.
\newblock In \emph{Proceedings of NAACL}.

\bibitem[{Fan et~al.(2017)Fan, Monti, Mathias, and Dreyer}]{fan2017transfer}
Xing Fan, Emilio Monti, Lambert Mathias, and Markus Dreyer. 2017.
\newblock Transfer learning for neural semantic parsing.
\newblock In \emph{Proceedings of the 2nd Workshop on Representation Learning
  for NLP}.

\bibitem[{Graves(2012)}]{graves2012supervised}
Alex Graves. 2012.
\newblock Supervised sequence labelling.
\newblock In \emph{Supervised sequence labelling with recurrent neural
  networks}, pages 5--13. Springer.

\bibitem[{Gulcehre et~al.(2016)Gulcehre, Ahn, Nallapati, Zhou, and
  Bengio}]{gulcehre2016pointing}
Caglar Gulcehre, Sungjin Ahn, Ramesh Nallapati, Bowen Zhou, and Yoshua Bengio.
  2016.
\newblock Pointing the unknown words.
\newblock \emph{Proceedings of ACL}.

\bibitem[{Gupta et~al.(2018{\natexlab{a}})Gupta, Rastogi, and
  Hakkani-Tur}]{gupta2018efficient}
Raghav Gupta, Abhinav Rastogi, and Dilek Hakkani-Tur. 2018{\natexlab{a}}.
\newblock An efficient approach to encoding context for spoken language
  understanding.
\newblock \emph{arXiv preprint arXiv:1807.00267}.

\bibitem[{Gupta et~al.(2018{\natexlab{b}})Gupta, Shah, Mohit, and
  Kumar}]{gupta}
Sonal Gupta, Rushin Shah, Mrinal Mohit, and Anuj Kumar. 2018{\natexlab{b}}.
\newblock Semantic parsing for task oriented dialog using hierarchical
  representations.
\newblock In \emph{Proceedings of EMNLP}.

\bibitem[{Gupta et~al.(2018{\natexlab{c}})Gupta, Shah, Mohit, Kumar, and
  Lewis}]{gupta2018semantic}
Sonal Gupta, Rushin Shah, Mrinal Mohit, Anuj Kumar, and Mike Lewis.
  2018{\natexlab{c}}.
\newblock Semantic parsing for task oriented dialog using hierarchical
  representations.
\newblock \emph{arXiv preprint arXiv:1810.07942}.

\bibitem[{Hakkani-T{\"u}r et~al.(2016)Hakkani-T{\"u}r, T{\"u}r, Celikyilmaz,
  Chen, Gao, Deng, and Wang}]{hakkani2016multi}
Dilek Hakkani-T{\"u}r, G{\"o}khan T{\"u}r, Asli Celikyilmaz, Yun-Nung Chen,
  Jianfeng Gao, Li~Deng, and Ye-Yi Wang. 2016.
\newblock Multi-domain joint semantic frame parsing using bi-directional
  rnn-lstm.
\newblock In \emph{Interspeech}, pages 715--719.

\bibitem[{Haklay and Weber(2008)}]{haklay2008openstreetmap}
Mordechai Haklay and Patrick Weber. 2008.
\newblock Openstreetmap: User-generated street maps.
\newblock \emph{Ieee Pervas Comput}, 7(4):12--18.

\bibitem[{Jia and Liang(2016)}]{jia2016data}
Robin Jia and Percy Liang. 2016.
\newblock Data recombination for neural semantic parsing.
\newblock In \emph{Proceedings of the 54th Annual Meeting of the Association
  for Computational Linguistics (Volume 1: Long Papers)}, volume~1, pages
  12--22.

\bibitem[{Kate et~al.(2005)Kate, Wong, and Mooney}]{kate2005learning}
Rohit~J Kate, Yuk~Wah Wong, and Raymond~J Mooney. 2005.
\newblock Learning to transform natural to formal languages.
\newblock In \emph{Proceedings of the National Conference on Artificial
  Intelligence}, volume~20, page 1062. Menlo Park, CA; Cambridge, MA; London;
  AAAI Press; MIT Press; 1999.

\bibitem[{Lawrence and Riezler(2016)}]{lawrence2016nlmaps}
Carolin Lawrence and Stefan Riezler. 2016.
\newblock Nlmaps: A natural language interface to query openstreetmap.
\newblock In \emph{Proceedings of COLING}.

\bibitem[{Liang(2013)}]{liang2013lambda}
Percy Liang. 2013.
\newblock Lambda dependency-based compositional semantics.
\newblock \emph{arXiv preprint arXiv:1309.4408}.

\bibitem[{Liang et~al.(2013)Liang, Jordan, and Klein}]{liang2013learning}
Percy Liang, Michael~I Jordan, and Dan Klein. 2013.
\newblock Learning dependency-based compositional semantics.
\newblock \emph{Computational Linguistics}, 39(2):389--446.

\bibitem[{Luong et~al.(2015)Luong, Le, Sutskever, Vinyals, and
  Kaiser}]{luong2015multi}
Minh-Thang Luong, Quoc~V Le, Ilya Sutskever, Oriol Vinyals, and Lukasz Kaiser.
  2015.
\newblock Multi-task sequence to sequence learning.
\newblock \emph{arXiv preprint arXiv:1511.06114}.

\bibitem[{Luong and Manning(2016)}]{luong2016achieving}
Minh-Thang Luong and Christopher~D Manning. 2016.
\newblock Achieving open vocabulary neural machine translation with hybrid
  word-character models.
\newblock In \emph{Proceedings of the ACL}.

\bibitem[{Mesnil et~al.(2015)Mesnil, Dauphin, Yao, Bengio, Deng, Hakkani-Tur,
  He, Heck, Tur, Yu et~al.}]{mesnil2015using}
Gr{\'e}goire Mesnil, Yann Dauphin, Kaisheng Yao, Yoshua Bengio, Li~Deng, Dilek
  Hakkani-Tur, Xiaodong He, Larry Heck, Gokhan Tur, Dong Yu, et~al. 2015.
\newblock Using recurrent neural networks for slot filling in spoken language
  understanding.
\newblock \emph{IEEE/ACM Transactions on Audio, Speech, and Language
  Processing}, 23(3):530--539.

\bibitem[{Meyers et~al.(2004)Meyers, Reeves, Macleod, Szekely, Zielinska,
  Young, and Grishman}]{meyers2004annotating}
Adam Meyers, Ruth Reeves, Catherine Macleod, Rachel Szekely, Veronika
  Zielinska, Brian Young, and Ralph Grishman. 2004.
\newblock Annotating noun argument structure for nombank.
\newblock In \emph{Proceedings of LREC}.

\bibitem[{Nivre(2003)}]{nivre2003efficient}
Joakim Nivre. 2003.
\newblock An efficient algorithm for projective dependency parsing.
\newblock In \emph{Proceedings of the 8th International Workshop on Parsing
  Technologies (IWPT}. Citeseer.

\bibitem[{Nivre(2004)}]{nivre2004incrementality}
Joakim Nivre. 2004.
\newblock Incrementality in deterministic dependency parsing.
\newblock In \emph{Proceedings of the Workshop on Incremental Parsing: Bringing
  Engineering and Cognition Together}, pages 50--57. Association for
  Computational Linguistics.

\bibitem[{Nivre(2008)}]{nivre2008algorithms}
Joakim Nivre. 2008.
\newblock Algorithms for deterministic incremental dependency parsing.
\newblock \emph{Computational Linguistics}, 34(4):513--553.

\bibitem[{Palmer et~al.(2005)Palmer, Gildea, and
  Kingsbury}]{palmer2005proposition}
Martha Palmer, Daniel Gildea, and Paul Kingsbury. 2005.
\newblock The proposition bank: An annotated corpus of semantic roles.
\newblock \emph{Computational linguistics}, 31(1):71--106.

\bibitem[{Ruder(2017)}]{ruder2017overview}
Sebastian Ruder. 2017.
\newblock An overview of multi-task learning in deep neural networks.
\newblock \emph{arXiv preprint arXiv:1706.05098}.

\bibitem[{Sha and Pereira(2003)}]{sha2003shallow}
Fei Sha and Fernando Pereira. 2003.
\newblock Shallow parsing with conditional random fields.
\newblock In \emph{Proceedings of the 2003 Conference of the North American
  Chapter of the Association for Computational Linguistics on Human Language
  Technology-Volume 1}, pages 134--141. Association for Computational
  Linguistics.

\bibitem[{Steedman(1996)}]{steedman1996surface}
Mark Steedman. 1996.
\newblock Surface structure and interpretation.

\bibitem[{Su and Yan(2017)}]{su2017crossdomain}
Yu~Su and Xifeng Yan. 2017.
\newblock Cross-domain semantic parsing via paraphrasing.
\newblock In \emph{Proceedings of EMNLP}.

\bibitem[{Thrun(1996)}]{thrun1996learning}
Sebastian Thrun. 1996.
\newblock Is learning the n-th thing any easier than learning the first?
\newblock In \emph{Proceedings of NIPS}.

\bibitem[{Vishal Ishwar~Naik(2018)}]{naik2018modal}
Rahul~Goel Vishal Ishwar~Naik, Angeliki~Metallinou. 2018.
\newblock Context aware conversational understanding for intelligent agents
  with a screen.

\bibitem[{Wang et~al.(2015{\natexlab{a}})Wang, Xue, and
  Pradhan}]{wang2015transition}
Chuan Wang, Nianwen Xue, and Sameer Pradhan. 2015{\natexlab{a}}.
\newblock A transition-based algorithm for amr parsing.
\newblock In \emph{Proceedings of NAACL}.

\bibitem[{Wang et~al.(2015{\natexlab{b}})Wang, Berant, and
  Liang}]{wang2015building}
Yushi Wang, Jonathan Berant, and Percy Liang. 2015{\natexlab{b}}.
\newblock Building a semantic parser overnight.
\newblock In \emph{Proceedings of ACL}.

\bibitem[{Zoph et~al.(2016)Zoph, Yuret, May, and Knight}]{zoph2016transfer}
Barret Zoph, Deniz Yuret, Jonathan May, and Kevin Knight. 2016.
\newblock Transfer learning for low-resource neural machine translation.
\newblock In \emph{Proceedings of EMNLP}.

\end{thebibliography}
\bibliographystyle{acl_natbib}
\end{document}